\documentclass{fcs}

\usepackage{amsmath}
\usepackage{bm}
\usepackage{multirow}
\usepackage{booktabs}
\usepackage{subcaption}
\usepackage{CJKutf8}
\usepackage{caption}
\usepackage{amssymb}
\usepackage{ulem}
\usepackage{pgfplots}
\definecolor{myred}{HTML}{E85642}
\definecolor{myblue}{HTML}{1280B0}

\volumn{ }
\doi{ }
\articletype{RESEARCH~ARTICLE}
\copynote{{\copyright} Higher Education Press and Springer-Verlag Berlin Heidelberg 2012}
\ratime{Received month dd, yyyy; accepted month dd, yyyy}
\email{bxwang2@iflytek.com}
\title{RE$^2$: improving Chinese grammatical error correction via retrieving appropriate examples with explanation}
\author{Baoxin WANG$^{1,2}$, Yumeng LUO$^3$, Yixuan WANG$^1$, Dayong WU$^2$, Wanxiang CHE$^1$, Shijin WANG$^2$}

\address{{1\quad Research Center for SCIR, Harbin Institute of Technology, Harbin 150001, China}\\
 {2\quad State Key Laboratory of Cognitive Intelligence, iFLYTEK Research, Hefei 230088, China}\\
 {3\quad Artificial Intelligence and Human Language Lab, Beijing Foreign Studies University, Beijing 100089, China}}

\begin{document}
\maketitle
\setcounter{page}{1}
\setlength{\baselineskip}{14pt}

\begin{abstract}
The primary objective of Chinese grammatical error correction (CGEC) is to detect and correct errors in Chinese sentences. Recent research shows that large language models (LLMs) have been applied to CGEC with significant results. For LLMs, selecting appropriate reference examples can help improve their performance. However, existing methods predominantly rely on text similarity for example retrieval, a strategy that frequently mismatches actual error patterns and retrieves lexically similar yet grammatically irrelevant sentences. To address this problem, we propose a method named RE$^2$, which \textbf{r}etrieves appropriate \textbf{e}xamples with \textbf{e}xplanations of grammatical errors. Instead of using text similarity of the input sentence, we use explanations of grammatical errors to select reference examples, which are used by LLMs to improve the performance of CGEC. We conduct experiments on two CGEC datasets and create a high-quality grammatical error explanation (GEE) dataset, which is not only used in our research but also serves as a valuable resource for future studies in both CGEC and GEE. The experimental results on the two datasets indicate that our proposed method effectively improves the performance of CGEC.
\end{abstract}

\Keywords{grammatical error correction, large language model, grammatical error explanation}

\section{Introduction}

Grammatical error correction (GEC) is a significant task in natural language processing, which aims to automatically identify and correct grammatical errors within a given sentence. It has important applications in areas such as education, news, and publishing. The predominant methods employed are sequence-to-sequence (Seq2Seq) \cite{zhao-etal-2019-improving, choe-etal-2019-neural, fang-etal-2023-improving} and sequence-to-edit (Seq2Edit) \cite{omelianchuk-etal-2020-gector, lai-etal-2022-type}. Recently, large language models (LLMs) have demonstrated remarkable efficacy across various natural language processing (NLP) tasks \cite{brown2020gpt, Zhao2023ASO, Ouyang2022instruct} and exhibited significant progress in the Chinese grammatical error correction (CGEC) task \cite{yang-quan-2024-alirector}.

\begin{figure}[t] 
\begin{center} 
	\includegraphics[width=8cm]{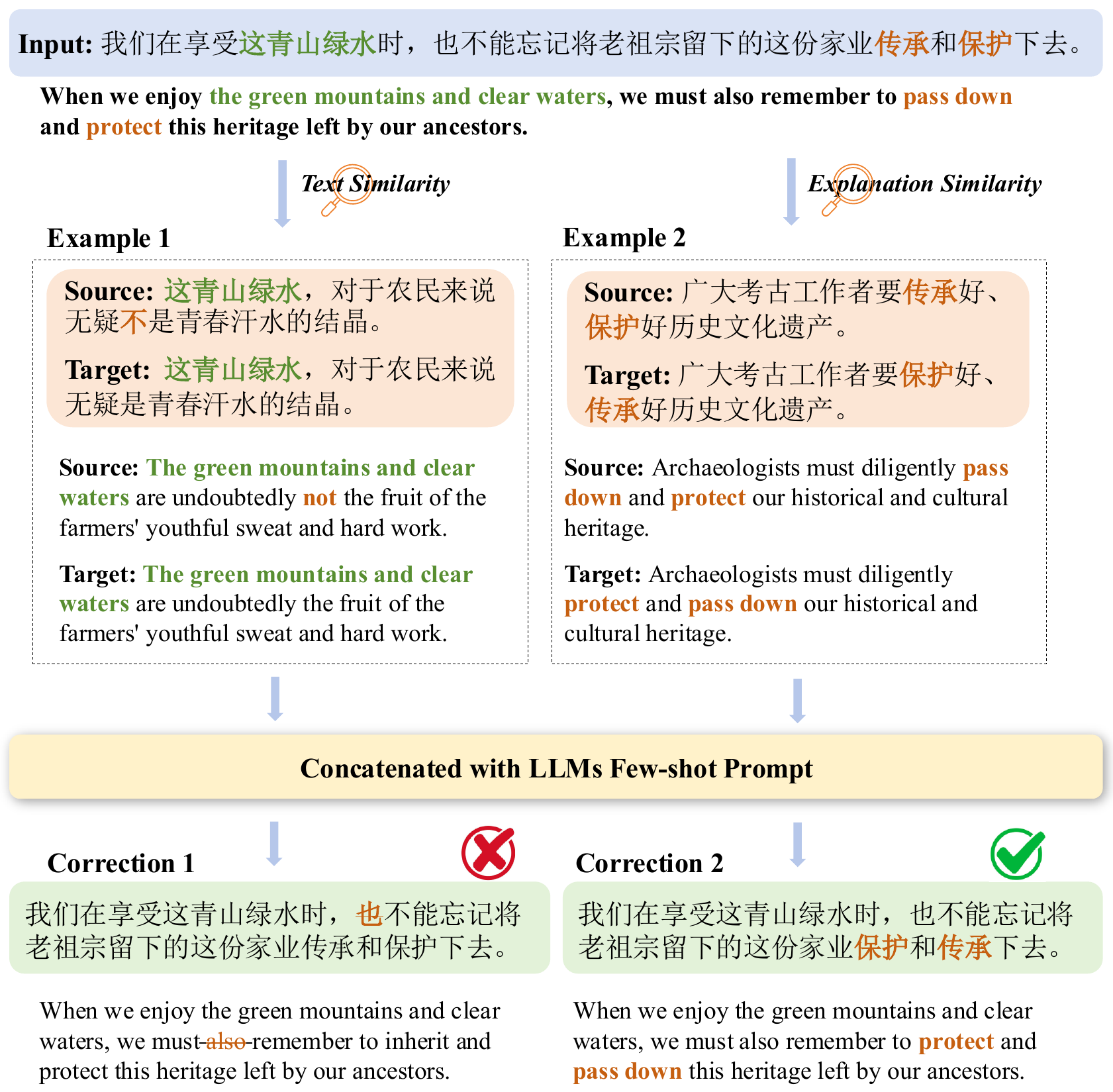}
	\caption{
	Examples of retrieving reference examples. Example 1 is retrieved using text similarity of the input sentence, while Example 2 is retrieved using the grammatical error explanation of the input sentence. The green-colored text is the overlapping part between the input and the text of Example 1, while the orange-colored text represents the corresponding errors and correction results. The overlapping part found between the reference example and the input using the text similarity method is “\begin{CJK*}{UTF8}{gbsn}这青山绿水\end{CJK*}” (the green mountain and clear waters), while the explanation similarity method can identify examples with the same errors in the order of “\begin{CJK*}{UTF8}{gbsn}传承\end{CJK*}” (pass down) and “\begin{CJK*}{UTF8}{gbsn}保护\end{CJK*}” (protect) 
	}
	\label{fig:examples} 
\end{center}
\end{figure}

Previous research has demonstrated that providing appropriate reference examples to models, whether decoder-only LLMs or encoder-decoder models, yields significant benefits on various tasks \cite{wang2022training, nashid2023code, huang2024raven}. We posit that appropriate examples would also prove beneficial for the CGEC task. Existing work primarily utilizes the approach of selecting examples that are similar to the input text, which suffers from two key limitations: (1) Sentences with similar surface structures may involve fundamentally different types of grammatical errors, leading to mismatched or irrelevant reference examples; (2) Text-similarity-based retrieval fails to account for the underlying error patterns or linguistic rules, limiting the model’s ability to generalize error-specific correction strategies. As shown in Figure \ref{fig:examples}, \emph{Example 1} is retrieved using the text similarity of the input sentence. Errors usually account for only a small part of the sentence, such as ``\begin{CJK*}{UTF8}{gbsn}传承\end{CJK*}'' (pass down) and ``\begin{CJK*}{UTF8}{gbsn}保护\end{CJK*}'' (protect) in the example. The overlapping part found between the reference example and the input using the text similarity method is “\begin{CJK*}{UTF8}{gbsn}这青山绿水\end{CJK*}” (the green mountain and clear waters). We can see that it is difficult to retrieve examples with similar errors directly through sentence similarity, making it challenging to provide effective guidance for LLMs.

Intuitively, examples with similar grammatical error patterns to the input sentence would be more helpful in assisting LLMs. Inspired by grammatical error explanation (GEE) \cite{fei-etal-2023-enhancing, song-etal-2024-gee}, we propose the RE$^2$ method, which uses grammatical error explanation similarity as the criterion for retrieving appropriate examples. Evidently, grammatical error explanations can better present the patterns of grammatical errors in sentences. As shown in Figure \ref{fig:examples}, the explanation similarity method can identify examples with the same errors in the order of “\begin{CJK*}{UTF8}{gbsn}传承\end{CJK*}” (pass down) and “\begin{CJK*}{UTF8}{gbsn}保护\end{CJK*}” (protect). By retrieving examples with similar explanations, the error patterns in the corresponding examples are also similar to those in the input sentence.

Based on the aforementioned ideas, we construct a high-quality dataset of Chinese grammatical error explanations. We utilize LLMs (GPT-4o) along with erroneous sentences, correct sentences, error types, and editing results to automatically generate explanations. To further improve the quality of explanations, we collect a set of official rough explanations from public school Chinese examinations and use LLMs to expand on these rough explanations. We perform a comprehensive expansion of the CGEC dataset FCGEC \cite{xu-etal-2022-fcgec}, resulting in the FCGEE (Fine-grained Chinese Grammatical Error Explanation) dataset.

In this paper, we utilize LLMs with in-context learning (ICL) \cite{rubin-etal-2022-learning} and supervised fine-tuning (SFT) to verify the effectiveness of \textbf{R}etrieving \textbf{E}xamples with similar \textbf{E}xplanation (RE$^2$). We first train a grammatical error explainer using the FCGEE dataset. Then, we use the explainer to automatically generate error explanations for input sentences, and subsequently retrieve similar examples through explanations. For the ICL method, these similar examples are inserted into the LLM's prompt as reference examples to assist in grammatical error correction. For the SFT method, during the training phase, we find $k$ examples with the most similar explanations for each training instance and concatenate these $k$ examples into the prompt as training data. The inference phase of SFT is the same as the ICL method.

We conducted experiments on two Chinese grammatical error correction datasets, FCGEC and NaCGEC \cite{ma2022linguistic}. The errors in these datasets are derived from primary and secondary school grammatical error questions for native Chinese speakers. These grammatical errors pose a certain level of difficulty even for native Chinese speakers, and error explanations are beneficial for understanding these grammatical errors. Our method shows significant improvement in both datasets, demonstrating its effectiveness. 

The main contributions of this paper are as follows:
\begin{itemize}
\item Construct a high-quality dataset of Chinese grammatical error explanations.
\item Propose the RE$^2$ method and explore the use of grammatical error explanations for retrieving appropriate reference examples.
\item Our method achieves the best performance on two datasets, and we conduct an in-depth analysis of grammatical error explanations.
\end{itemize}

\begin{table*}[htbp]
\centering
\caption{Examples of different error types from the FCGEC dataset} 
\renewcommand\arraystretch{1.3}
\scalebox{0.9}{
\begin{tabular}{ll}
\toprule
\textbf{Error Type} & \textbf{Sentence} \\ \hline
\multirow{3}{*}{Incorrect Word Order} & \begin{CJK*}{UTF8}{gbsn}任何一种文明的发展都是与其他文明碰撞、\sout{融合、交流}(交流、融合)的过程\end{CJK*}\\ 
& The development of any civilization is a process of collision, \sout{fusion and exchange}\\
&(exchange and fusion) with others. \\\cline{1-2}
\multirow{3}{*}{Incorrect Word Collocation} & \begin{CJK*}{UTF8}{gbsn}为了提高这次舞会的档次\sout{和规模}，举办方特邀中国人民解放军乐团现场演奏。\end{CJK*}\\ 
&In order to improve the grade \sout{and scale} of this dance, the organizer specially invited\\
&the Chinese People's Liberation Army Band to perform live. \\\cline{1-2}
\multirow{3}{*}{Component Missing} & \begin{CJK*}{UTF8}{gbsn}这种治疗方法具有见效快，无副作用(的特点)，以达到标本兼治的目的。\end{CJK*}\\ 
&This treatment method has (the feature of) quick effect and no side effects, so as to \\
&achieve the purpose of treating both the symptoms and the root causes. \\\cline{1-2}
 \multirow{2}{*}{Component Redundancy} & \begin{CJK*}{UTF8}{gbsn}有一部分网友却对雷锋及雷锋精神提出了各种各样的所谓\sout{质疑}(疑问)\end{CJK*}\\ 
&Some netizens have raised various so-called questions about Lei Feng and his spirit.\\\cline{1-2}
\multirow{3}{*}{Structure Confusion} & \begin{CJK*}{UTF8}{gbsn}由于资金不足\sout{的限制}，学校计划修建的图书楼和医疗室只好暂缓施工。\end{CJK*}\\ 
&Due to \sout{limit of} insufficient funds, the school's planned library building and medical \\
&room had to be postponed.\\\cline{1-2}
\multirow{2}{*}{Ambiguity} & \begin{CJK*}{UTF8}{gbsn}山上的水宝贵，我们把它留给\sout{晚上来}的人喝。\end{CJK*}\\ 
&The water is precious, we leave it to people who come to drink \sout{at night} (late). \\\cline{1-2}
\multirow{4}{*}{Illogical} & \begin{CJK*}{UTF8}{gbsn}如今的手机已不再是单纯的通信工具，\sout{因而}(还)成为人们生活中的贴身伴侣，用\end{CJK*}\\ 
&\begin{CJK*}{UTF8}{gbsn}来尽情表现个人品位。\end{CJK*}\\
&Today's mobile phone is no longer a simple communication tool, \sout{so} it has become \\
&a personal companion in people's life, used to express personal taste to the fullest.\\
\bottomrule
\end{tabular}}
\label{all_dataset}
\end{table*}

\begin{CJK*}{UTF8}{gbsn}
\begin{table*}[t]\small
\newcommand{\tabincell}[2]{\begin{tabular}{@{}#1@{}}#2\end{tabular}}
\begin{center}
\caption{Prompt template of grammatical error explanation generation for GPT-4o with edits and error types. The error types come from the FCGEC dataset}
\scalebox{0.8}{
\begin{tabular}{ll}
\specialrule{0.05em}{3pt}{7pt}
& This is a Chinese Grammatical Error Explanation task. Based on a pair of incorrect and correct sentences, the edits, and the type of \\
& grammatical error, you are required to generate an explanation according to the given template. The edits are the changes needed to \\
& correct the incorrect sentence. \\
& \\
& All error types: \\
& IWC (Incorrect Word Collocation): a word-level grammatical error in which the related words are combined in the improper pattern \\
& CM (Component Missing): some components (e.g., subject and object) of the sentence are missing \\
& CR (Component Redundancy): some components are redundant \\
& SC (Structure Confusion): a syntax-level grammatical error that combines two similar grammatical structures into a single incorrect one \\
& IWO (Incorrect Word Order): incorrect word order in word-level or pragmatic-level \\
& ILL (Illogical): logical error \\
& AM (Ambiguity): ambiguity caused by unclear references or expressions \\
& \\
& Example: \\
& Source: The development of science and technology, the revitalization of the economy, and even the progress of the entire society, \\
& the main reason is determined by the high quality of the workforce and the cultivation of a large number of outstanding talents. \\
& Target: The development of science and technology, the revitalization of the economy, and even the progress of the entire society, is \\
& determined by the high quality of the workforce and the cultivation of a large number of outstanding talents. \\
& Edits: [24, ``the main reason'', ``''] \\
& Error Type: SC \\
& Explanation step by step: The sentence contains a Structural Confusion (SC) error because the two similar structures, ``the main \\
& reason is'' and ``is determined by'' are mixed together, leading to unclear sentence structure. The phrase ``the main reason'' should be \\
& removed to eliminate the confusion and make the sentence more concise and clear. \\
& \\
& Now, please generate an error explanation according to the given sentence pair, edits, and error type, following the format described above. \\
& \{input sentence\} \\
& \{output sentence\} \\
& Edits: \{edits\} \\
& Error Type: \{error type\} \\
& Explanation step by step: \\
\specialrule{0.05em}{7pt}{3pt}
\end{tabular}}
\label{tab:prompt_with_edits}
\end{center}
\end{table*}
\end{CJK*}

\section{Related Work}
\subsection{CGEC Methods}
Existing CGEC approaches can be categorized into three mainstream paradigms: Seq2Seq, Seq2Edit, and LLMs. Seq2Seq method typically uses a standard encoder-decoder architecture of Transformer \cite{transformer2017vaswani}, which is an end-to-end approach treating CGEC as a neural machine translation task \cite{fu2018youdao,zhu2020incorporating,wang-etal-2024-lm}. Rather than generating a completely new sentence, Seq2Edit methods regard CGEC as a process of token-level sequence labeling, predicting a sequence of edit operations, such as insertion, deletion, and substitution \cite{malmi-etal-2019-encode,awasthi-etal-2019-parallel,omelianchuk-etal-2020-gector}. Recently, research has been extended to explore LLMs’ performance on the CGEC task.

\subsection{GEC with Large Language Models}
\noindent\textbf{In-Context Learning (ICL)}\;
Previous studies assessed the GEC performance of closed-source models by carefully designing prompts with in-context learning strategies \cite{fang2023chatgpt,li2023effectiveness}. The results showed that while closed-source LLMs, like ChatGPT, demonstrate excellence in error detection and generation fluency, they exhibit more over-correction and fall far short of the performance of the baseline small models.

\noindent\textbf{Supervised Fine-tuning (SFT)}\;
Some recent works have studied the potential of open-source LLMs architectures on the CGEC task through supervised fine-tuning (SFT), and reveal that it helps to enhance model performance. GrammarGPT \cite{fan2023grammargpt} performed instruction fine-tuning for an open-source LLM and significantly surpassed the existing SOTA baseline while requiring less training data. Yang and Quan \cite{yang-quan-2024-alirector} proposed Alirector (alignment-enhanced corrector) by utilizing SFT as well as knowledge distillation, effectively mitigating the over-correction issue of decoder-only LLMs.

\noindent\textbf{Grammatical Error Explanation (GEE)}\;
In GEE tasks, LLMs are prompted to explain reasons for corrections in natural language for pairs of ungrammatical and grammatical sentences. Besides, studies have demonstrated that explicitly providing edits in prompts can guide LLMs to cover all correction points and thus yield marked improvements in the models' explanation outputs \cite{song-etal-2024-gee,kaneko2023controlled}. To better evaluate the explanations that LLMs generated, Kaneko and Okazak \cite{kaneko2023controlled} annotated several existing GEC datasets and built the XGEC dataset, in which each data includes an incorrect text, a correct text, and explanations for each edit. Our experiments reveal that directly generating error explanations \cite{song-etal-2024-gee} yields relatively low accuracy. To address this limitation, we propose a rough explanation guided method to enhance the performance of grammatical error explanation.

\noindent\textbf{Retrieval-based Prompt Engineering}\;
Retrieval-based prompt engineering is to retrieve the most related labeled instances and then concatenate them into the input as an in-context demonstration to feed into the model, which helps optimize the prompt to improve few-shot inference \cite{li2021prefix}. Wang et al. \cite{wang2022training} proposed a method to retrieve from the labeled training data, which significantly improves baseline performance on various NLU and NLG tasks. Nashid et al. \cite{nashid2023code} introduce CEDAR, an automatic prompt creation method that retrieves similar code demonstrations to enhance few-shot learning, which outperforms task-specific and fine-tuned models in test assertion generation and program repair tasks.

Within the above literature, the question of how to select appropriate reference examples to improve LLM performance on challenging CGEC tasks remains inadequately addressed. Previous reference example selection methods for CGEC tasks often struggle to accurately identify suitable reference examples. To tackle this challenge, we propose a systematic framework aimed at retrieving contextually relevant examples, accompanied by grammatical error explanations.

\begin{figure}[t] 
\begin{center} 
	\includegraphics[width=7.5cm]{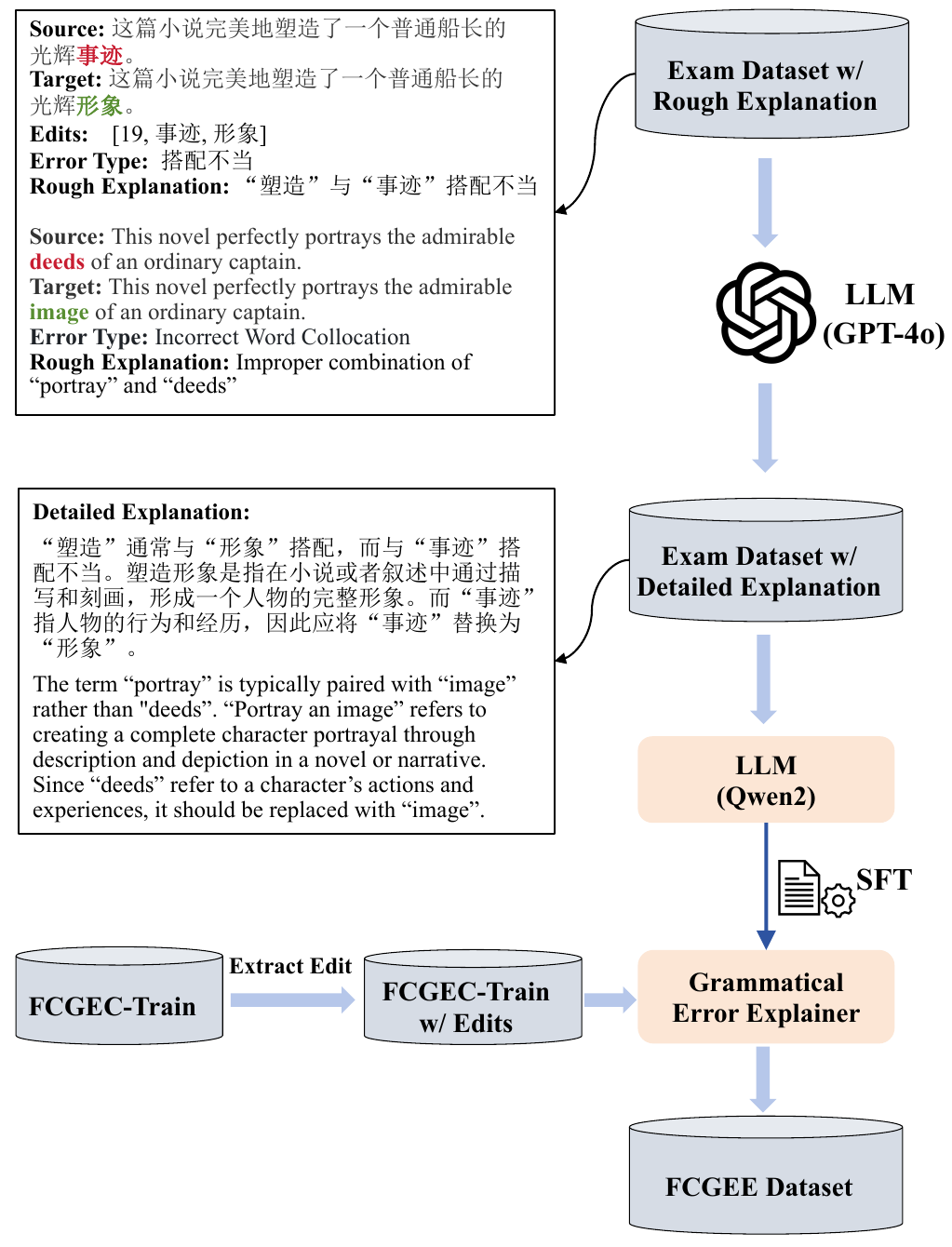}
	\caption{
 	Rough explanation guided method for constructing the grammatical error explanation dataset FCGEE. The red-colored text indicates the incorrect part, and the green-colored text indicates the corrected result
	}
	\label{fig:explain_gen} 
\end{center}
\end{figure}

\section{Methodology}
\subsection{Task Definition}
The Chinese grammatical error correction task aims to detect and correct grammatical errors in text. Given an input text sequence $X=\{x_1, x_2, ..., x_n\}$, the goal of the CGEC task is to automatically identify grammatical errors within the text and generate a corrected target sequence $Y=\{y_1, y_2, ..., y_m\}$. Xu et al. \cite{xu-etal-2022-fcgec} construct a Fine-Grained Chinese Grammatical Error Correction (FCGEC) dataset from public school Chinese examinations for native students. The FCGEC dataset categorizes grammatical errors into seven types: Incorrect Word Order (IWO), Incorrect Word Collocation (IWC), Component Missing (CM), Component Redundancy (CR), Structure Confusion (SC), Illogical (ILL), and Ambiguity (AM). Specific examples corresponding to these seven error types are presented in Table \ref{all_dataset}.

\subsection{Grammatical Error Explanation Dataset}
To achieve more accurate example selection, we experiment with two approaches for constructing a grammatical error explanation dataset. The first approach relies entirely on automated methods using LLMs to generate error explanations. Similar to \cite{song-etal-2024-gee}, we first use the LTP tool \cite{che-etal-2010-ltp} to perform word segmentation on the sentences, and then use difflib library to extract edit results based on the source and target sentences. We then input the source sentence, target sentence, edits, and error type (included in the FCGEC dataset) into LLMs to generate the explanation. We use an example to assist the LLM in generating explanations that meet our requirements, which include error type, error cause, and modification suggestions. We test multiple types of prompts and ultimately use the one-shot prompt template shown in Table \ref{tab:prompt_with_edits} to generate grammatical error explanations via GPT-4o.

According to the evaluation in Section 3.3, the accuracy of explanations generated by the above method with GPT-4o is only 71\%. This is likely due to the high difficulty of errors in the FCGEC dataset, which consists of examination questions designed for native Chinese speakers, requiring deep understanding of Chinese text to provide correct explanations. 

To improve the quality of explanations, we propose a rough explanation guided method for generating grammatical error explanations. We collect sentences with official rough error explanations from primary and secondary school exams, ultimately obtaining 10,598 instances with rough explanations. As illustrated in Figure \ref{fig:explain_gen}, the explanations provided with the examination questions are relatively rough, and many errors require deeper understanding and analysis. We propose using the rough explanations as guidance to assist LLMs in generating more in-depth explanations. The prompt template is similar to the prompt with edits in Table \ref{tab:prompt_with_edits}, except for adding a rough explanation to the input. Using this prompt, GPT-4o generates high-quality grammatical error explanations. In the FCGEC dataset, the majority of examples contain only one erroneous edit. For cases involving multiple error edits, we employ the same processing method as that used for a single edit.

\begin{figure}[t] 
\begin{center} 
	\includegraphics[width=5cm]{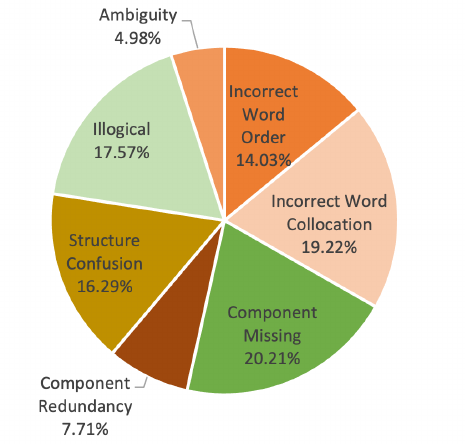}
	\caption{
        The error type distribution of the FCGEE dataset
	}
	\label{fig:type_dist} 
\end{center}
\end{figure}

\begin{figure*}[t] 
\begin{center} 
	\includegraphics[width=14cm]{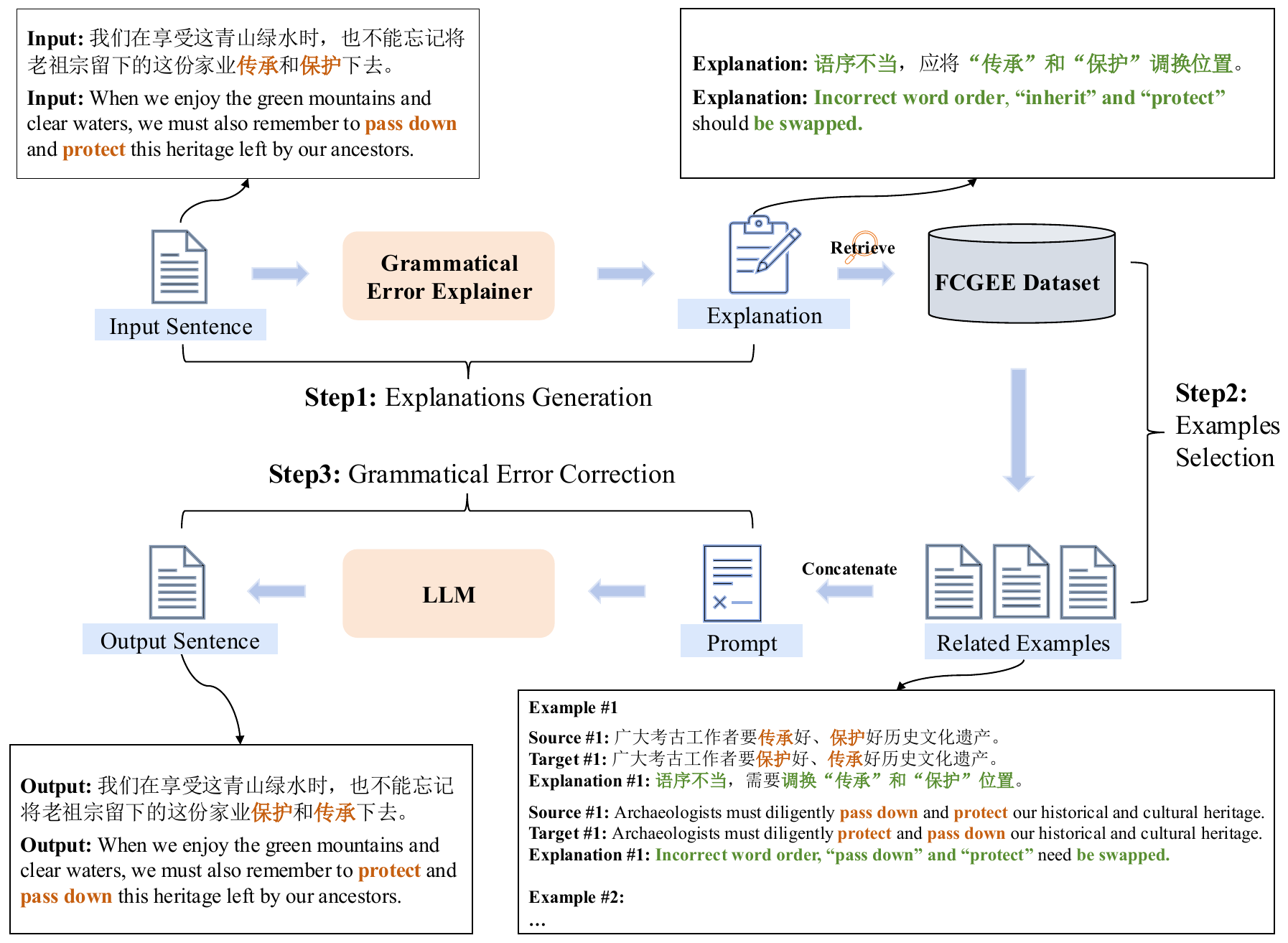}
	\caption{
        Overview of RE$^2$. The prompt template is shown in Table \ref{tab:gec_prompt_templates}. The green-colored text represents the overlapping part of the explanation texts, while the orange-colored text corresponds to the error and correction parts in the example
	}
	\label{fig:train_proc} 
\end{center}
\end{figure*}

The generation process of the grammatical error explanation dataset is shown in Figure \ref{fig:explain_gen}. We use the exam dataset with detailed explanations as training data to train an LLM for generating explanations. This model takes the erroneous sentence, correct sentence, edits, and error type as input, and outputs the detailed grammatical error explanation. By training on this data using the open-source LLM Qwen2-7B-Instruct \cite{qwen2}, we obtain an explanation generation model that works with known editing results. We still use LTP and difflib here to extract editing results based on the source and target sentences for the FCGEC dataset. Using the explanation generation model, we perform inference on the FCGEC training data with edits to obtain the grammatical error explanations dataset, which we call FCGEE. The explanation dataset contains only sentences with grammatical errors from the FCGEC dataset, totaling 19,758 sentences. The error type distribution of the FCGEE dataset is shown in Figure \ref{fig:type_dist}, where component missing accounts for the highest proportion at 20.21\%, while ambiguity has the lowest proportion at only 4.98\%. The detailed evaluation of explanation quality is introduced in Section 3.3.

\subsection{Retrieving Examples with Explanation (RE$^2$)}

Appropriate reference examples can help LLMs improve performance. In this section, we introduce in-context learning and supervised fine-tuning to verify the effectiveness of retrieving examples with similar explanations for CGEC. All the examples mentioned in this paper refer to those containing the source sentence and the target sentence, not including error explanations. The overview of RE$^2$ is illustrated in Figure \ref{fig:train_proc}. We first generate explanations of the grammatical errors based on the input sentence. Then, we retrieve the $k$ most similar reference examples from the FCGEE dataset based on the explanation. Finally, these $k$ reference examples are provided to the LLM to assist it in correcting the grammatical errors.

\noindent\textbf{Explanations Generation}\;
To produce high-quality grammatical error explanations for input sentences, we train a grammatical error explainer based on Qwen2-7B-Instruct. The training data is derived from the FCGEE dataset constructed in the preceding subsection. The explainer takes as input only the input sentence without error types or editing results. By feeding the input sentence into the explainer, we obtain potential explanations for grammatical errors. Notably, even if the input sentence is error-free, the explainer is designed to generate hypothetical error explanations, identifying potential positions for refinement within the sentence.

\noindent\textbf{Examples Selection}
The similarity of the grammatical error explanations can effectively reflect the similarity of grammatical errors. Therefore, we calculate the similarity between the explanations of the input sentence and those of each example in the FCGEE dataset and retrieve the similar examples with the $k$ highest explanation similarity. Similar to \cite{Tang2024FsPONERFP}, we adopt the TF-IDF cosine similarity to measure the similarity between the input explanation and the explanations from the FCGEE dataset.

\begin{CJK*}{UTF8}{gbsn}
\begin{table*}[t]\small
\begin{center}
\centering
\caption{Prompt template of RE$^2$ with examples and without examples}
\scalebox{0.9}{
\begin{tabular}{l}
\specialrule{0.05em}{3pt}{3pt}
\multicolumn{1}{c}{\textbf{Prompt with GEC examples}} \\
\specialrule{0.05em}{3pt}{5pt}
下面给出中文语法纠错参考示例。 \\
原句：\{Source \#1\} 纠正后：\{Target \#1\} \\
原句：\{Source \#2\} 纠正后：\{Target \#2\} \\
...... \\
原句：\{Source \#k\} 纠正后：\{Target \#k\} \\
参考上述纠错示例，请判断下面句子是否有语法错误，如果有请直接进行纠正，如果没有请直接输出原句。 \\
\{Input\} \\
\\
\textbf{Translation:} \\
Here is a reference example for Chinese grammar correction: \\
Original sentence: \{Source \#1\} Corrected sentence: \{Target \#1\} \\
Original sentence: \{Source \#2\} Corrected sentence: \{Target \#2\} \\
...... \\
Original sentence: \{Source \#k\} Corrected sentence: \{Target \#k\} \\
Referring to the above error correction example, please check if there are any grammatical errors in the following \\
sentence. If there are, please correct them directly. \\
If not, please output the original sentence directly: \\
\{Input\} \\
\specialrule{0.05em}{5pt}{3pt}
\multicolumn{1}{c}{\textbf{Prompt without GEC examples}} \\
\specialrule{0.05em}{3pt}{5pt}
请判断下面句子是否有语法错误，如果有请直接进行纠正，如果没有请直接输出原句： \\
\{Input\} \\
\\
\textbf{Translation:} \\
Please check if there are any grammatical errors in the following sentence. If there are, please correct them directly. \\
If not, please output the original sentence directly: \\
\{Input\} \\
\specialrule{0.05em}{5pt}{3pt}
\end{tabular}}
\label{tab:gec_prompt_templates}
\end{center}
\end{table*}
\end{CJK*} 

\noindent\textbf{Grammatical Error Correction}
Providing appropriate examples can effectively enhance the grammatical error correction capabilities of LLMs. For the in-context learning approach, the k most similar examples are incorporated into the prompt shown in Table \ref{tab:gec_prompt_templates}. Then, the LLM directly predicts results based on the input prompt.

For supervised fine-tuning, the reference examples should be included in the training data. During the training phase, drawing on the work of \cite{wang2022training}, we retrieve the $k$ most similar reference examples from the FCGEE dataset for each training instance and integrate these reference examples into the prompt the same as the ICL method to form the input, with the output being the correct sentence. Through this procedure, we assemble a set of training data that contains reference examples. To improve the model's correction ability in scenarios without reference examples, we also construct an equivalent amount of training data that does not include reference examples, with the specific prompt as shown in Table \ref{tab:gec_prompt_templates}. The training data, which includes both reference and non-reference samples, is used to train the Qwen-7B model.

During the inference phase, the $k$ most similar examples are retrieved based on the explanation. If the similarity of the most similar example is below a threshold $\theta$, indicating that suitable reference examples are not found within the training set, these similar examples will be disregarded, and the sentence will be corrected using the prompt without examples. If the highest similarity score is greater than or equal to the threshold $\theta$, the $k$ examples will serve as reference examples for the LLM. 

\section{Experiments}
\subsection{Experimental Settings}
\textbf{Dataset}\;
We focus exclusively on native speaker datasets when conducting experiments. Compared to second-language learner datasets, native speaker datasets include grammatical errors with a higher level of difficulty and complexity, making them more challenging for LLMs. For the training set, we use the training set of FCGEC, which primarily consists of erroneous sentences from public school Chinese examinations for native students from elementary to high school and correct sentences from news websites. The FCGEC dataset includes a training set, validation set, and test set, with sizes of 36,340, 2,000, and 3,000 respectively. We extend it by adding the grammatical error explanation for each erroneous sentence, providing support for subsequent research on GEE and GEC. 

Considering that the references of FCGEC-Test have not been open-sourced, which can only be evaluated by submitting results online, we also utilize the NaCGEC dataset as our test set. The NaCGEC dataset, also derived from native student examinations, employs a nearly identical error type framework to FCGEC. We remove the overlapping sentences between the NaCGEC test set and the FCGEC training set to prevent data leakage, leaving 3,900 sentences in the remaining NaCGEC test set.

\noindent\textbf{Hyperparameter Settings}\;
We employ Qwen2-7B-Instruct as the base model for training, with a learning rate of $1\times10^{-5}$, and 3 epochs of training. The training is conducted in a full parameter fine-tuning manner. We train Qwen2-7B-Instruct using LLaMA-Factory \cite{zheng2024llamafactory} on eight Nvidia A40 48GB GPUs. During the inference phase, we use beam search for inference, setting the beam size to 8 and the temperature to 1. The explainer and GEC models use the same training and inference settings. We find that the performance shows no significant improvement when the number of examples $k$ exceeds 3, so $k$ is set to 3. The similarity threshold $\theta$ is set to 0.6. We use the character-level precision, recall, and $F_{0.5}$ scores adopted by previous researchers using the ChERRANT scorer \cite{zhang-etal-2022-mucgec} as our evaluation metrics. 

We employ LTP \cite{che-etal-2010-ltp} for word segmentation and utilize the TfidfVectorizer from the scikit-learn library to compute the TF-IDF vectors for sentence pairs. The n-gram range is configured from 2 to 3, while the document frequency (DF) is statistically calculated using the FCGEE training dataset. The cosine similarity metric is then applied to measure the similarity between sentences.

\noindent\textbf{Baseline Methods}\;
We select multiple state-of-the-art models as our baseline models, including the Seq2Seq models, the Seq2Edit models, and LLMs. The specific baselines are as follows:
\begin{itemize}
\item \textbf{LaserTagger} \cite{malmi-etal-2019-encode}: A text editing model that reconstructs the output by applying a set of edit operations on the input sequence.
\item \textbf{PIE} \cite{awasthi-etal-2019-parallel}: A new parallel iterative editing architecture can generate outputs in parallel, which greatly reduces the decoding latency for long input sequences.
\item \textbf{GECToR} \cite{omelianchuk-etal-2020-gector}: A novel text editing approach is proposed with three training stages: pre-training on synthetic data, fine-tuning on erroneous parallel corpus and fine-tuning on a combination of erroneous and non-erroneous parallel corpus.

\item \textbf{STG} \cite{xu-etal-2022-fcgec}: The STG model is designed to correct grammatical errors in low-resource settings through a decomposition into three modules: Switch, Tagger, and Generator, with a focus on non-autoregressive approaches to enhance efficiency.

\item \textbf{LM-Combiner} \cite{wang-etal-2024-lm}: The LM-Combiner is a rewriting model designed to directly modify over-corrections in GEC system outputs for Chinese texts, which enables it to generate correct sentences by combining original and over-corrected texts.

\item \textbf{Alirector} \cite{yang-quan-2024-alirector}: Alirector is an alignment-enhanced corrector designed to address overcorrection issues in decoder-only LLMs by iteratively refining corrections through an alignment process and transferring alignment knowledge to improve overall CGEC performance.

\begin{table}[!tp]
    \centering
    \caption{
    Experimental results of our method on the FCGEC test set.
    The first group indicates baseline models, the second group indicates GPT-4o with ICL, and the third group indicates Qwen2-7B with SFT. The p-value obtained from the t-test is 0.003, which is highly significant (p < 0.01)
    }
    \scalebox{1.}{
    \begin{tabular}{lcccccc}
    \toprule
         \multirow{2}{*}{\textbf{Method}} & \multicolumn{3}{c}{\textbf{FCGEC-test}}  \\
         \cmidrule(lr){2-4}
        & $\textbf{P}$ & $\textbf{R}$ & $\textbf{F}_{0.5}$ \\
        \midrule
        LaserTagger  & 36.60 & 31.16 & 35.36\\
        PIE & 29.15 & 29.77 & 29.27\\
        GECToR (Chinese) & 30.68 & 21.65 & 28.32\\
        STG  & 48.19 & 37.14 & 45.48 & \\
        BART-Chinese-large & 37.49 & 38.87 & 37.76 \\
        GPT2-medium & 56.71 & 24.79 & 45.10\\
        LM-Combiner & 55.67 & 39.04 & 51.30\\
        Alirector & 64.49 & 36.22 & 55.78 \\
        \midrule
        GPT-4o Zero-Shot & 29.03 & 20.17 & 26.69 \\
        GPT-4o Few-Shot & 29.16 & 18.27 & 26.05 \\
        GPT-4o TextSim & 40.04 & 29.88 & 37.49 \\
        GPT-4o RE$^2$ & 49.22 & 41.34 & 47.41 \\
        \midrule
        Qwen2-7B & 65.57 & 39.50 & 57.92 \\
        Qwen2-7B RE$^2$ & \textbf{66.33} & \textbf{42.80} & \textbf{59.76} \\
    \bottomrule
    \end{tabular}
    }
    \label{tab:main_experiment}
\end{table}

\item \textbf{Zero-Shot}: This method does not apply reference examples and directly uses the prompt without GEC examples as shown in Table \ref{tab:gec_prompt_templates}.

\item \textbf{Few-Shot}: This method randomly selects three examples as references and uses the prompt with GEC examples in Table \ref{tab:gec_prompt_templates}.

\item \textbf{TextSim}: This method selects the three examples from the FCGEC dataset with the highest text similarity to the input sentence as reference examples and uses the prompt template the same as Few-Shot.

\end{itemize}

\begin{CJK*}{UTF8}{gbsn}
\begin{table*}[!tp]\small
\begin{center}
\caption{Examples of grammatical error explanations. The source sentence incorrectly pairs \emph{production} with \emph{supply} and \emph{sell}. It should pair \emph{vegetables} with \emph{supply} and \emph{sell}. The orange-colored text in the source and target sentences corresponds to the erroneous and corrected parts, respectively. The red-colored text indicates problematic explanations of GPT-4o, while the blue-colored text provides the corresponding correct explanation from our explainer} 
\scalebox{0.74}{
\begin{tabular}{ll}
\toprule
\multirow{3}{*}{\textbf{Source}}        & \textcolor{orange}{该基地每年的无公害蔬菜的生产量}除供应本省主要市场外，还销往河南、河北等省。 \\
& \textcolor{orange}{The annual production of pollution-free vegetables} in this base is not only supplied to the main markets in the province, but also sold \\
& to provinces such as Henan and Hebei.\\
\midrule
\multirow{3}{*}{\textbf{Target}}        & \textcolor{orange}{该基地每年生产的无公害蔬菜}除供应本省主要市场外，还销往河南、河北等省。 \\
& \textcolor{orange}{The pollution-free vegetables produced by this base every year} are not only supplied to the main markets in the province, but also \\
& sold to provinces such as Henan and Hebei. \\
\midrule[0.8pt]
\multirow{6}{*}{\textbf{GPT-4o}} & 该句存在词语搭配错误，因为\textcolor{myred}{原句中的“每年的无公害蔬菜的生产量”搭配不当}。“生产量”与“每年”和“无公害蔬菜”之间的关\\
& 系应该是“每年生产的无公害蔬菜”，这样表达更为准确。因此应将“每年的无公害蔬菜的生产量”改为“每年生产的无公害蔬\\
& 菜”，以确保词语搭配正确。\\
& There is a word collocation error in this sentence because \textcolor{myred}{the original sentence's \emph{annual production of pollution-free vegetables} is not} \\
& \textcolor{myred}{properly matched}. The relationship between \emph{production}, \emph{annual} and \emph{pollution-free vegetables} should be expressed as \emph{pollution-free} \\
& \emph{vegetables produced annually}, which is more accurate. Therefore, the term \emph{annual production of pollution-free vegetables} should be \\
& changed to \emph{annual production of pollution-free vegetables} to ensure correct word matching.\\
\midrule
\multirow{5}{*}{\textbf{Our Explainer}} & \textcolor{myblue}{句子中的“无公害蔬菜的生产量”与“供应”、“销往”等动词搭配不当}，因为“生产量”指的是产量的多少，而不是具体的蔬菜。\\
& 正确的表达应该是“生产的无公害蔬菜”，这样才能与后面的“供应”、“销往”等动词搭配得当。 \\
& \textcolor{myblue}{The phrase \emph{production of pollution-free vegetables} in the sentence is not paired properly with verbs such as \emph{supply} and \emph{sell}}, as \\ 
& \emph{production} refers to the amount of output, rather than specific vegetables. The correct expression should be \emph{produced pollution-free} \\
& \emph{vegetables}, so that it can be paired properly with verbs such as \emph{supply} and \emph{sell}. \\
\bottomrule
\end{tabular}}
\label{tab:explain_examples}
\end{center}
\end{table*}
\end{CJK*} 

\begin{table}[t]
    \centering
    \caption{
    Experimental results on the NaCGEC test set
    }
    \scalebox{1.}{
    \begin{tabular}{lcccccc}
    \toprule
         \multirow{2}{*}{\textbf{Method}} & \multicolumn{3}{c}{\textbf{NaCGEC-test}}  \\
         \cmidrule(lr){2-4}
        & $\textbf{P}$ & $\textbf{R}$ & $\textbf{F}_{0.5}$ \\
        \midrule
        BART-Chinese-large & 34.14 & 33.66 & 34.04 \\
        STG  & 46.63 & 36.24 & 44.10 \\
        \midrule
        GPT-4o Zero-Shot & 26.17 & 18.42 & 24.14 \\
        GPT-4o Few-Shot & 27.28 & 17.91 & 24.69 \\
        GPT-4o TextSim & 40.87 & 33.24 & 39.07 \\
        GPT-4o RE$^2$ & 45.59 & \textbf{40.18} & 44.39 \\
        \midrule
        Qwen2-7B & 54.69 & 36.27 & 49.65 \\
        Qwen2-7B RE$^2$ & \textbf{57.05} & 37.49 & \textbf{51.66} \\
    \bottomrule
    \end{tabular}
    }
    \label{tab:main_experiment_table_nacgec}
\end{table}

\subsection{Main Results}

Table \ref{tab:main_experiment} presents the experimental results on the FCGEC dataset. For in-context learning, we evaluate the performance of the large language model GPT-4o using four distinct example selection methods: Zero-Shot, Few-Shot, TextSim, and RE$^2$. Selecting examples based on text similarity yields a performance improvement of over 10 points compared to random selection. Furthermore, employing the RE$^2$ method, which retrieves examples with the $k$ highest explanation similarity, enhances the TextSim method by additional 9.92\%. These results suggest that the RE$^2$ method effectively retrieves appropriate reference examples, thereby augmenting the LLM's grammatical error correction capabilities.

For the supervised fine-tuning methods, LLMs have achieved superior performance on the CGEC task. Alirector utilizes the Baichuan2-7B model for training and significantly outperforms previous Seq2Seq and Seq2Edit models. Moreover, the large language model Qwen2-7B demonstrates even greater potential. Our proposed RE$^2$ method, leveraging the Qwen2-7B model, employs reference examples based on explanation similarity to improve CGEC performance. RE$^2$ method further enhances the performance of the Qwen2-7B model by 1.84\%, resulting in the best performance.
\begin{table}[tp]
    \centering
    \caption{
    The accuracy for different grammatical error explanation methods. Sugg. refers to the modification suggestion
    }
    \scalebox{0.88}{
    \begin{tabular}{lccc}
    \toprule
         \textbf{Methods} & \textbf{Type (\%)} & \textbf{Cause (\%)} & \textbf{Sugg. (\%)}\\
        \midrule
        GPT-4o w/ edits & 100 & 71 & 92 \\
        Explainer w/ edits & 100 & 88 & 96 \\
        Explainer w/o edits & 76 & 73 & 68 \\
        GPT-4o w/ rough expl. & 100 & 97 & 99 \\
    \bottomrule
    \end{tabular}
    }
    \label{explanation_quality_table}
\end{table}

\begin{CJK*}{UTF8}{gbsn}
\begin{table*}[!tp]\small
\renewcommand\arraystretch{1.2}
\begin{center}
\caption{Examples of grammatical error explanations for sentences without errors}
\scalebox{0.75}{
\begin{tabular}{ll}
\toprule
\multirow{3}{*}{\textbf{Example \#1}}        & \textbf{Source}: 我省出版界积极组织精品图书参展，介绍、展现本省出版界面对加入WTO的新形势加快产业化发展的新思路、新风貌。 \\
& Our province’s publishing sector actively organizes exhibitions of high-quality books to introduce and demonstrate its new ideas and new \\
& styles in accelerating industrial development in response to  the new situation of joining the WTO. \\
& \textbf{Explanation}: 句子中的“介绍”与“新思路、新风貌”搭配不当。“介绍”通常与“情况”、“作品”等相搭配，而“展现”更适合与“风貌”、\\
& “思路”等抽象概念搭配。因此应将“介绍”改为“展现”以保持搭配的正确性。\\
& The word \textit{introduce} in this sentence is inappropriately paired with \textit{new ideas and new styles}. \textit{Introduce} is generally used in combination \\
& with concrete concepts such as \textit{situations} or \textit{works}, while \textit{demonstrate} is more appropriate when used with abstract concepts like \textit{styles} or \\
& \textit{ideas}. Therefore, \textit{introduce} should be replaced with \textit{demonstrate} to ensure proper word pairing. \\
\midrule
\multirow{3}{*}{\textbf{Example \#2}}        & \textbf{Source}: 据了解，我省汉简发现早、数量多、研究历史长、在海内外影响大。 \\
& It is understood that the Han bamboo slips in our province were discovered at an early stage. They have a large quantity and a long research \\
& history, and hold significant influence both domestically and internationally. \\
& \textbf{Explanation}: 句子中的“发现早、数量多、研究历史长、在海内外影响大”是对“我省汉简”的描述，但是“研究历史长”后缺少了\\
& “的”字，使得句子结构不完整，导致成分残缺。因此，需要在“研究历史长”后添加“的”字，使句子结构完整，表达清晰。\\
& In this sentence, the phrase \textit{discovered at an early stage} and \textit{have a large quantity and a long research history, and hold significant influence} \\
& \textit{both domestically and internationally} are used to describe our province's Han Dynasty bamboo slips. However, there is a missing structural \\
& particle "的(of)" after \textit{a long research history}, which makes the sentence structure incomplete and leads to a Component Missing error. \\
& Therefore, it is necessary to add a particle "的" after \textit{a long research history} to complete the sentence structure and make the expression clear. \\
\midrule
\multirow{6}{*}{\textbf{Example \#3}} & \textbf{Source}: 世界各国人民创造的灿烂文化，是人类共同的宝贵财富，我们应该通过交流互鉴和创造性发展，使它焕发出新的生命力。\\
& The splendid culture created by people all over the world is the common treasure of mankind. We should infuse it with new vitality through \\
& exchanges, mutual learning and creative development. \\
& \textbf{Explanation}: 句子中的“通过交流互鉴和创造性发展，使它焕发出新的生命力”缺少主语，导致句子结构不完整。为了使句子完整\\
& 且明确，需要在“通过交流互鉴和创造性发展，使它焕发出新的生命力”前添加主语“我们”。\\
& The phrase \textit{infuse it with new vitality through exchanges, mutual learning and creative development} lacks a subject, resulting in an \\
& incomplete sentence structure. To make the sentence complete and clear, we need to add a subject \emph{we} before \emph{infuse it with new vitality} \\
& \emph{through exchanges, mutual learning and creative development}. \\
\bottomrule
\end{tabular}}
\label{tab:explain_examples_wo_errors}
\end{center}
\end{table*}
\end{CJK*}

\begin{table}
    \centering
    \caption{
    Experimental results of different example selection methods. ``w/ examples'' refers to the use of reference examples for all instances regardless of explanation similarity. ``Oracle'' refers to the selection of reference examples based on correct explanations
    }
    \scalebox{0.9}{
    \begin{tabular}{lccc}
    \toprule
         \multirow{2}{*}{\textbf{Method}} & \multicolumn{3}{c}{\textbf{FCGEC-dev}} \\
         \cmidrule(lr){2-4}
        & $\textbf{P}$ & $\textbf{R}$ & $\textbf{F}_{0.5}$\\
        \midrule
        Qwen2-7B & 51.81 & 38.87 & 48.57 \\
        Qwen2-7B TextSim & 50.42 & 37.24 & 47.09 \\
        \midrule
        Qwen2-7B RE$^2$  & 54.50 & 39.49 & 50.65 \\
        \;\;\;\;\; w/ examples  & 53.56 & 39.72 & 50.07 \\
        \;\;\;\;\; Oracle & \textbf{55.25} & \textbf{42.05} & \textbf{51.99} \\
    \bottomrule
    \end{tabular}
    }
    \label{reference_example_table}
\end{table} 

Table \ref{tab:main_experiment_table_nacgec} presents the experimental results for NaCGEC, aligning with the findings from FCGEC. The RE$^2$ method surpasses other example selection methods and effectively enhances the grammatical error correction capabilities of LLMs.

\subsection{Explanation Quality Analysis}

Table \ref{tab:explain_examples} presents specific examples generated by the explainer. GPT-4o's explanations are based directly on correction results without deep analysis. This may be because GPT-4o lacks specialized training for correcting complex Chinese grammatical errors. Even with a few reference examples, the model struggles to accurately provide the causes of errors. In contrast, our explainer clearly understands that the error is due to improper collocation, provides a detailed analysis, and correctly gives the modification suggestion. This demonstrates that rough explanations offer LLMs a more specific direction for explanations, enabling them to produce higher-quality error explanations.

Two master's degree candidates who are native Chinese speakers manually evaluate the quality of the generated grammatical error explanations, primarily assessing the error type, error cause, and modification suggestions. For human evaluation, we randomly select a modest sample of 100 instances, with the evaluation results shown in Table \ref{explanation_quality_table}. Because we provide the error types and editing results to GPT-4o, it achieves relatively good performance for the error type and modification suggestions. However, GPT-4o's accuracy in explaining error causes is only 71\%. For our specially trained explainer based on editing results, the error cause accuracy improves to 88\%. The explainer based solely on input sentences achieves 73\%, still higher than the GPT-4o model provided with editing results, which also demonstrates the high quality of our explainer.

Table \ref{tab:explain_examples_wo_errors} shows explanations generated for sentences without errors. In Example 1, the explainer mistakenly identifies a collocation error due to the pairing of two verbs and two nouns, which can be prone to misinterpretation. In Examples 2 and 3, the absence of errors leads the explainer to produce irrelevant explanations. We observe that in such cases, the generated explanations frequently struggle to match high-similarity explanations in FCGEE. Although there are no actual errors, the highlighted potential error positions are common error spots. This information is helpful for retrieving appropriate reference examples.

\subsection{Impact of Example Selection and Explanation Quality}
We conduct experiments on different methods of selecting reference examples for LLMs with SFT. The experimental results are shown in Table \ref{reference_example_table}. We can observe that the method of directly using text similarity to find reference examples performs even worse than directly training the model without examples. This indicates that reference examples retrieved based on text similarity are not appropriate and instead introduce more noise that affects the model's performance for SFT methods. On the other hand, reference examples retrieved through grammatical error explanations clearly perform better than the model without examples, demonstrating the effectiveness of the RE$^2$ method. The model using golden explanations achieves the best results, which also suggests that if we can continue to improve the effect of explanation generation, it can help further enhance performance.

\begin{table*}[!tp]
    \centering
    \caption{
    The ROUGE-L scores corresponding to different decoding parameters of the explainer
    }
    \scalebox{0.9}{
    \begin{tabular}{lccc}
    \toprule
         \textbf{Decoding Parameters} & \textbf{ROUGE-L (\%)}\\
        \midrule
        Oracle & 100 \\
        sample=false, temperature=1, beam size=8 & 75.61 \\
        sample=false, temperature=1, beam size=6 & 70.22 \\
        sample=false, temperature=1, beam size=4 & 68.46 \\
        sample=false, temperature=1, beam size=2 & 63.04 \\
        sample=false, temperature=1, beam size=1 & 58.08 \\
        sample=true, temperature=0.9, top\_k=5, top\_p=0.75 & 55.60 \\
        sample=true, temperature=0.9, top\_k=12, top\_p=0.95 & 53.50 \\
    \bottomrule
    \end{tabular}
    }
    \label{tab:decoding_parameters}
\end{table*}

\begin{figure}[t] 
\begin{center} 
	\includegraphics[width=7.7cm]{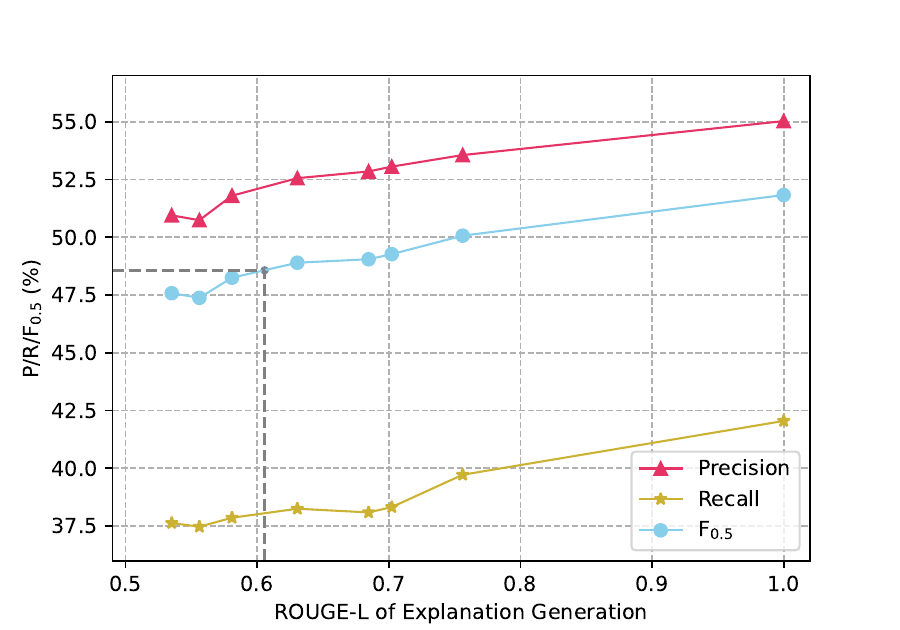}
	\caption{
	The results of our method with different explanation ROUGE-L scores on the FCGEC-dev dataset. We generate explanations with different ROUGE-L scores via the explainer using different decoding parameters. The dashed line corresponds to using Qwen2-7B without reference examples
	}
	\label{fig:explanation_quality} 
\end{center}
\end{figure}

\begin{figure}[t] 
\begin{center} 
	\includegraphics[width=7.7cm]{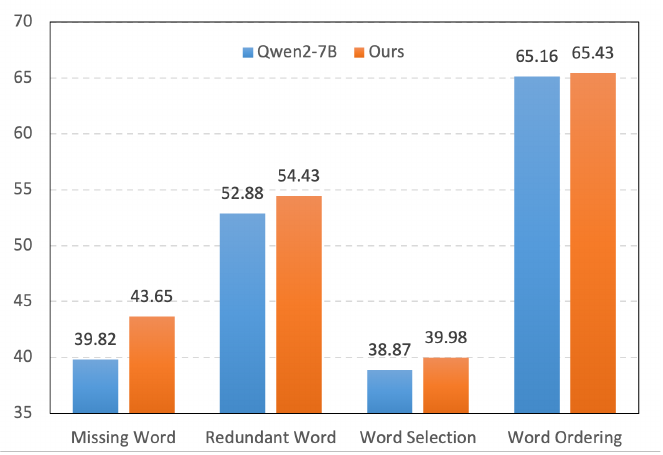}
	\caption{
	The results of $F_{0.5}$ for different error types, including missing (M), redundant (R), word selection (S), and word-order (W), on the FCGEC-dev set
	}
	\label{fig:error_type} 
\end{center}
\end{figure}

We verify the influence of explanation quality on the performance of RE$^2$. Utilizing different decoding parameters, we generate explanations for grammatical errors that exhibit different levels of quality, which are subsequently evaluated using the ROUGE-L metric \cite{lin-2004-rouge}. Table \ref{tab:decoding_parameters} shows the ROUGE-L scores corresponding to the explanations generated with different decoding parameters of the explainer. Using beam search can effectively enhance the performance of the explainer, while employing sampling methods leads to a decrease in the quality of the generated explanations, which affects the grammatical error correction performance of the LLMs.

Figure \ref{fig:explanation_quality} shows the influence of explanation quality on experimental results. When the ROUGE-L score of generated explanations is greater than 0.6, our proposed method can effectively improve the performance. When the ROUGE-L score is less than this value, low-quality explanations will bring negative effects on model performance by selecting inappropriate reference examples. With the increase in explanation quality, there is still room for further improvement in the effectiveness of our method.

\subsection{Analysis for Error Types and Similarity Threshold}
As illustrated in the Figure \ref{fig:error_type}, after adopting the RE$^2$ method, there is a significant improvement in missing errors, with an increase of 3.83\%. This indicates that providing some appropriate similar reference examples helps the model better add corresponding tokens to sentences with missing elements. These errors are relatively difficult for LLMs to learn directly, but through reference examples, it becomes easier for the model to learn this type of error. Redundancy and substitution errors also show notable improvements. However, the improvement in word order errors is not very significant. This might be because there are various situations of disordered sequences, making it difficult for reference examples to provide much effective help.

\begin{figure}[t] 
\begin{center} 
	\includegraphics[width=7.7cm]{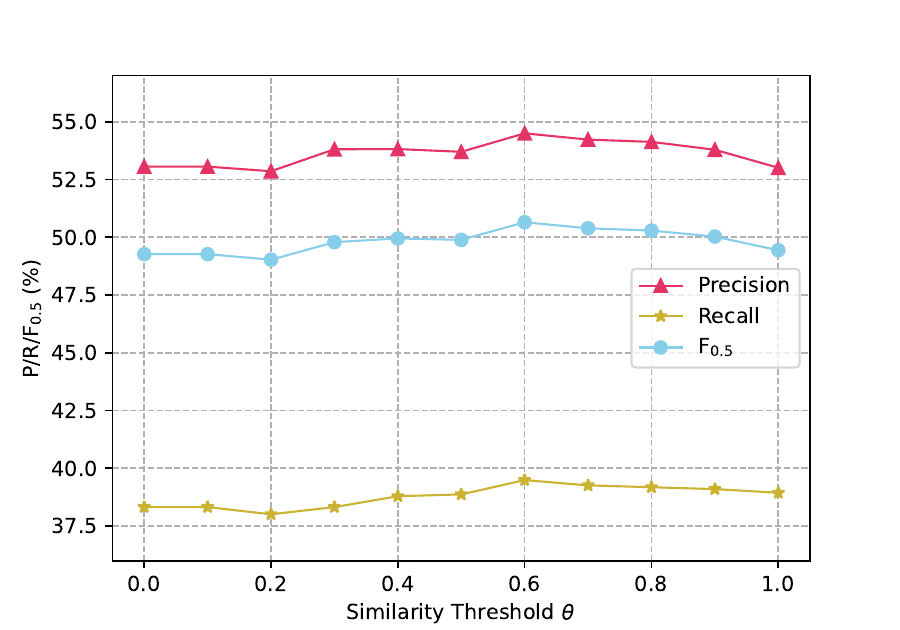}
	\caption{
	The experimental results of explanation similarity thresholds on the FCGEC-dev dataset
	}
	\label{fig:explanation_threshold} 
\end{center}
\end{figure}

\begin{table}
    \centering
    \caption{
    Experimental results of different retrieval methods on the FCGEC-dev dataset
    }
    \scalebox{0.9}{
    \begin{tabular}{lccc}
    \toprule
         \multirow{2}{*}{\textbf{Method}} & \multicolumn{3}{c}{\textbf{FCGEC-dev}} \\
         \cmidrule(lr){2-4}
        & $\textbf{P}$ & $\textbf{R}$ & $\textbf{F}_{0.5}$ \\
        \midrule
        Qwen2-7B & 51.81 & 38.87 & 48.57 \\
        Qwen2-7B RE$^2$ w/ BGE & 54.88 & 38.96 & 50.73 \\
        Qwen2-7B RE$^2$ w/ BM25 & 54.22 & 39.38 & 50.42 \\
        Qwen2-7B RE$^2$ w/ TF-IDF & 54.50 & 39.49 & 50.65 \\
    \bottomrule
    \end{tabular}
    }
    \label{retrieval_method_table}
\end{table} 

We analyze the impact of the similarity threshold on the model's performance. As shown in Figure \ref{fig:explanation_threshold}, the performance of CGEC is influenced by the explanation similarity threshold $\theta$. The overall trend indicates that as the threshold $\theta$ increases, the performance initially improves, but when the threshold continues to rise, the performance gradually declines. The F$_{0.5}$ score performs best on the validation set when the threshold $\theta$ is set to 0.6. Therefore, in the main experiments, we set the threshold to 0.6.

\subsection{Comparative Analysis of Retrieval Methods}
To determine which retrieval method to use, we conduct an ablation study analysis. Specifically, we experiment with TF-IDF, BM25 \cite{bm25stephen}, and sentence-embedding retriever (BGE-zh-base \cite{bge2024xiao}). The experimental results are shown in Table \ref{retrieval_method_table}, where it can be observed that the TF-IDF method outperforms BM25 and achieves comparable performance to the BGE-zh-base retriever. This may be due to the fact that the BGE-zh-base model captures the semantic information of explanations but lacks the ability to represent and capture the specific types and causes of errors within the explanations, resulting in only marginal improvements in example selection performance. Moreover, BGE-zh-base retriever consumes significantly more computational resources. In a CPU environment with Intel(R) Xeon(R) Gold 6240 CPU, generating a vector takes 875 ms, which is much slower compared to TF-IDF and BM25 methods (around 50 ms). Considering these factors, we use TF-IDF for retrieval in this work. In future research, we aim to explore stronger vector representation models that are specifically fine-tuned for this task.

\begin{CJK*}{UTF8}{gbsn}
\begin{table*}[t]\small
\renewcommand\arraystretch{1.2}
\begin{center}
    \caption{Examples of explanations for Chinese spelling and grammatical errors} 
\scalebox{0.9}{
\begin{tabular}{ll}
\toprule
\multirow{5}{*}{\textbf{SIGHAN \#1}}        & \textbf{Source}: 真麻烦你了。希望你们好好的\textbf{跳无}。\\
& \textbf{Target}: 真麻烦你了。希望你们好好的\textbf{跳舞}。 \\
& \textbf{Explanation}: “跳无”是一个字形错误。“无”和“舞”在发音上相近，但含义完全不同。“无” \\
& 表示“没有”，在语义上无法与“跳”搭配；而“舞”表示“舞蹈”，与“跳”搭配后语义通顺。\\
& 因此，应将“无”改为“舞”，使句子表达准确。\\
\midrule
\multirow{5}{*}{\textbf{SIGHAN \#2}}        & \textbf{Source}: 不好意思，我\textbf{咪路}，可不可以帮我？ \\
& \textbf{Target}: 不好意思，我\textbf{迷路}，可不可以帮我？ \\
& \textbf{Explanation}: “咪路”是一个字形错误。“咪”和“迷”在发音上相似，但含义不同。“咪”通\\
& 常用于拟声词或与猫有关的词语，而“迷”在这里表示“迷失、走失”，与“路”搭配后语义\\ 
& 通顺。因此，应将“咪”改为“迷”，使句子表达正确无误。\\
\midrule
\multirow{4}{*}{\textbf{CGED \#1}} & \textbf{Source}: 但是同时，这种说法是从\textbf{人间}出现在地球上，一直存在的说法吧。\\
& \textbf{Target}: 但是同时，这种说法是从\textbf{人类}出现在地球上，一直存在的说法吧。\\
& \textbf{Explanation}: 用词不当，“人间”是指人类生活的地方或世间，而这里指的是“人类”的出现。\\
& “人类”更加贴合语境和逻辑表达。\\
\midrule
\multirow{5}{*}{\textbf{CGED \#2}} & \textbf{Source}: 上面的方式已经没用了。\textbf{人间}生活下去的时候，必须要水。 \\
& \textbf{Target}: 上面的方式已经没用了。\textbf{人类}生活下去的时候，必须要水。\\
& \textbf{Explanation}: 用词不当，在句子“人间生活下去的时候”中，“人间”指的是世间或人类生存\\
& 的环境，而这里实际要表达的是“人类”这一主体，强调需要水来维持人类的生存。因此，\\
& 应将“人间”修改为“人类”，以使表达准确且符合语境。 \\
\bottomrule
\end{tabular}}
\label{tab:explain_examples_cged}
\end{center}
\end{table*}
\end{CJK*}

\begin{table*}[t]
    \centering
    \caption{
    The experimental results of the RE$^2$ method on the CGED 2018 test set
    }
    \scalebox{1.}{
    \begin{tabular}{lcccccc}
    \toprule
         \multirow{2}{*}{\textbf{Method}} & \multicolumn{3}{c}{\textbf{Sentence-Level}} & \multicolumn{3}{c}{\textbf{Position-Level}} \\
         \cmidrule(lr){2-7}
        & $\rm \textbf{Precision}$ & $\rm \textbf{Recall}$ & $\rm \textbf{F1}$ & $\rm \textbf{Precision}$ & $\rm \textbf{Recall}$ & $\rm \textbf{F1}$ \\
        \midrule
        Qwen2-7B  & 66.96 & 91.23 & 77.23 & 25.67 & 25.08 & 25.37 \\
        Qwen2-7B TextSim & 69.23 & 90.18 & 78.33 & 28.22 & 24.70 & 26.34 \\ 
        Qwen2-7B RE$^2$ & \textbf{70.18} & \textbf{92.35} & \textbf{79.75} & \textbf{28.72} & \textbf{27.48} & \textbf{28.09} \\
    \bottomrule
    \end{tabular}
    }
    \label{result_cged}
\end{table*}

\section{Discussion on the Limitations and Scope of Application}
\subsection{Inference Speed}
The RE$^2$ method, in contrast to directly leveraging large language models (LLMs) for CGEC, incorporates two additional steps: generating explanations for grammatical errors and retrieving relevant examples. Consequently, this approach incurs greater time overhead compared to other methods. Experimental results indicate that the grammatical error correction time for the RE$^2$ method is approximately 2 times longer than that of directly employing LLMs for error correction. Although the generation of explanations extends the overall correction time, the generated grammatical error explanations significantly aid language learners in understanding their mistakes. Exploring methods to accelerate the correction process remains an important direction for future research.

\subsection{Retrieval Method}
Drawing on previous research works, we employ the TF-IDF method to enhance the retrieval speed. Nevertheless, there are still numerous other sentence representation retrieval methods grounded in neural networks. These methods have the potential to retrieve more precise reference examples, thereby achieving superior results. We consider exploring more powerful vector representation models specifically trained for this task as part of our future research work.

\subsection{Scope of Application}
Chinese spelling errors in the SIGHAN dataset \cite{tseng-etal-2015-introduction} are predominantly attributed to phonetic and graphemic similarities. In this study, we attempt to provide explanations for these errors by adopting an approach similar to the RE$^2$ method. Specifically, we leverage the GPT-4o model, coupled with carefully designed prompts, to generate detailed explanations for the observed spelling errors. For instance, the first two rows of Table \ref{tab:explain_examples_cged} illustrate representative explanations for Chinese spelling errors in the SIGHAN dataset.

Chinese spelling errors often arise from incorrect input via Pinyin input methods. Such errors can occur in any character within a sentence, and the generated explanations for different errors tend to exhibit significant overlap, offering limited specificity or variation. This lack of differentiation poses challenges in identifying suitable reference examples based on the explanations. As a result, Chinese spelling correction (CSC) tasks dominated by phonetic and graphemic errors are not well-suited for the RE$^2$ method.

Based on the above analysis, the RE$^2$ method is not suitable for tasks involving irregular errors caused by carelessness. In contrast, the Chinese Grammatical Error Diagnosis (CGED) dataset \cite{rao2018overview}, which contains grammatical errors made by non-native learners, includes fewer careless errors and more errors caused by language differences. The last two rows of Table \ref{tab:explain_examples_cged} provide examples of grammatical error explanations from the CGED dataset. For instance, the term "\begin{CJK*}{UTF8}{gbsn}人间\end{CJK*}" (world) in Japanese is synonymous with the Chinese term "\begin{CJK*}{UTF8}{gbsn}人类\end{CJK*}" (human), so Japanese native speakers often confuse the two terms. In this section, we conduct experiments on the CGED dataset to verify the effectiveness of the RE$^2$ method with supervised fine-tuning. The CGED 2017 training set is used for training, and the CGED 2018 test set is used for evaluation. The evaluation employs the sentence-level and position-level error detection methods \cite{rao2018overview}. The experimental results are shown in Table \ref{result_cged}. The model using the RE$^2$ method performs significantly better than approaches relying solely on text similarity or those that do not use reference examples. This further demonstrates that the RE$^2$ method is more suitable for tasks where the input text exhibits certain inherent patterns.

\section{Conclusion}
In this paper, we first generate a high-quality grammatical error explanation dataset with the help of official rough explanations and LLMs. Furthermore, utilizing the grammatical error explanation dataset, we propose the RE$^2$ method which retrieves appropriate reference examples through grammatical error explanation, which can effectively improve the effect of LLMs and achieves the best performance on both FCGEC and NaCGEC datasets. In future work, more methods that utilize the grammatical error explanation dataset can be explored to further enhance the performance of GEC.

\section*{Acknowledgment}
    We thank all reviewers and  Xiaoxue Wang for their constructive comments.
    We gratefully acknowledge the support of the National Natural Science Foundation of China (NSFC) via grant 62236004, 62206078 and 62476073.

    \bibliographystyle{fcs}
    \bibliography{fcs}

\end{document}